\documentclass{article}
\usepackage{spconf,amsmath,amssymb,graphicx,hyperref}
\usepackage{booktabs}

\title{LiDAR Signals for Robot Locomotion: Spherical Range Images Enable Robust Sim-to-Real Transfer}
\name{
Nian Liu$^{1,\dagger}$ \qquad
Xinyi Tong$^{2,\dagger}$ \qquad
Ming Ren$^{1}$ \qquad
Long Ye$^{2*}$ \qquad
Zhaofeng He$^{1*}$
}

\address{
$^{1}$ School of Artificial Intelligence, Beijing University of Posts and Telecommunications, Beijing, China \\
$^{2}$ School of Data Science and Intelligent Media, Communication University of China, Beijing, China \\
$^{\dagger}$ These authors contributed equally. \qquad
$^{*}$ Corresponding authors.
}
\begin{document}
\ninept
\maketitle
\begin{abstract}
Existing legged-locomotion systems commonly rely on depth images or online elevation maps. However, depth-based policies suffer from perceptual sim-to-real discrepancies due to imperfect depth sensor modeling, while map-based approaches are vulnerable to drift and misalignment. LiDAR offers direct measurements of the surrounding geometry, but raw LiDAR scans lack a compact, structured representation for high-frequency control. We propose a LiDAR perception pipeline that maintains perceptual consistency across simulation and reality through sensing- and representation-level alignment. At the sensor level, we recover frame-wise angular sampling patterns from the physical LiDAR and replay them in simulation with GPU ray casting. At the representation level, simulated and real returns undergo the same spherical range image encoding, yielding a fixed-size policy observation for end-to-end training and direct deployment without teacher--student distillation, online elevation mapping, or adaptation. Experiments show that our range-image policies achieve stronger sim-to-real transfer than depth- and elevation-map-based baselines, while ablations confirm the importance of the proposed sensing components. More visualizations and real-world videos are available at the \href{https://lidar-img.github.io/}{\underline{\textit{https://lidar-img.github.io}}}.
\end{abstract}
\section{Introduction}
Learning-based control has enabled legged robots to acquire increasingly agile locomotion skills \cite{lee2020learning,hoeller2024anymal}. Extending these capabilities to challenging terrain, however, relies on a perception representation that faithfully represents the geometry and remains consistent between simulation and real-world deployment.

Existing methods primarily use depth images \cite{agarwal2023legged,zhuang2023robot,cheng2024extreme,luo2024pie,gadde2025blindspots,rudin2026parkour} or online elevation maps \cite{fankhauser2018robust,miki2022learning,hoeller2024anymal,sun2025perceptive,dong2025marg}. Many depth-based methods distill a teacher with access to a privileged local height map into a student conditioned on onboard depth \cite{agarwal2023legged,zhuang2023robot,cheng2024extreme,zhuang2025humanoid}, yet this supervision is ambiguous when terrain configurations requiring different actions are indistinguishable in the student's partial view \cite{chanesane2025solo}. More fundamentally, simulators render clean depth, whereas real sensors exhibit illumination- and motion-dependent artifacts, including missing values, boundary errors, and jitter. Generic noise models cannot fully reproduce these artifacts \cite{zhuang2023robot,duan2024learning,sun2026dpl,sun2026now}, creating a sim-to-real mismatch that can distort terrain perception and degrade policy performance on hardware. LiDAR provides direct range measurements that are less sensitive to illumination than camera-based depth. Although raw scans can be processed by dedicated point-cloud encoders \cite{qi2017pointnet,nahrendra2024dreamwaqpp,wang2025omni}, many legged locomotion systems instead rely on an additional online mapping pipeline that uses pose-based registration, temporal fusion, and filtering to construct robot-centric elevation maps \cite{fankhauser2018probabilistic,miki2022elevation}. The accuracy of these maps, however, depends critically on the pose estimates used to register successive measurements. Odometry or SLAM drift can therefore introduce gaps, spurious steps, and spatial misalignment with the physical terrain \cite{miki2022elevation}, compromising control decisions that depend on accurate local geometry.

In this paper, we convert LiDAR signals into structured spherical range images that directly condition a locomotion policy. To improve sim-to-real transfer, we address perceptual discrepancies by aligning the sensing process and observation representation across simulation and deployment. At the sensor level, we recover frame-wise angular scan patterns from the physical LiDAR and replay them in simulation through parallel ray casting. Articulated robot geometry is included to model self-occlusion. At the representation level, simulated and real returns undergo the same gravity-aligned temporal aggregation and rasterization onto a fixed angular grid. This shared encoding preserves angular organization and local range structure while providing a compact, fixed-size policy input in both domains. The resulting pipeline supports end-to-end policy training in simulation and direct deployment without teacher--student distillation, online elevation mapping, or post-training adaptation.

We compare our range-image policy with alternative perception designs, such as depth images and elevation maps, on challenging terrains. Our method outperforms these alternatives by a substantial margin in sim-to-real traversal success. Ablations assess the proposed sensing and representation design.

\begin{figure*}[t]
    \centering
    \includegraphics[width=\textwidth]{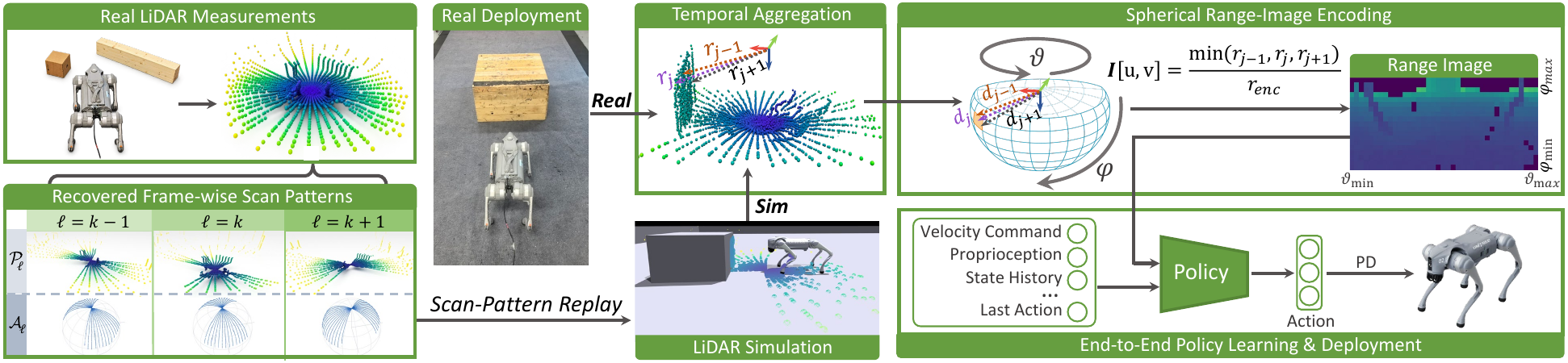}
    \caption{Overview of our pipeline. Real-world scan patterns are replayed in simulation, while simulated and real LiDAR returns share the same temporal aggregation and spherical range-image encoding for end-to-end policy learning and deployment.}
    \label{fig:method}
    \vspace{-1em}
\end{figure*}

\section{Method}

\subsection{Problem Formulation}
We formulate terrain locomotion as a partially observable Markov decision process and learn a stochastic policy $\pi_{\theta}$ through reinforcement learning. At each control step $t$, the policy receives an observation $\mathbf{o}_t$ and samples an action $\mathbf{a}_t\sim\pi_{\theta}(\cdot\mid\mathbf{o}_t)$ for execution by the low-level controller. The observation components are specified with the policy architecture in Sec.~\ref{sec:policy}. Using rewards that promote command tracking and stable terrain traversal, we optimize the policy by maximizing the expected discounted return
\begin{equation}
    \theta^{\star}
    =
    \arg\max_{\theta}
    \mathbb{E}_{\tau\sim p_{\theta}}
    \left[
        \sum_{t=0}^{T-1}\gamma^{t}r_t
    \right],
\end{equation}
where $p_{\theta}(\tau)$ is the policy-induced trajectory distribution, $T$ the horizon, $r_t$ the reward, and $\gamma\in[0,1)$ the discount factor.
\subsection{Sensor-Level LiDAR Simulation}
Faithful LiDAR simulation requires both accurate scene geometry and a sensor model that reproduces how the physical sensor samples that geometry, particularly for non-repetitive LiDARs whose ray directions vary across scans. We therefore extract a temporally ordered sequence of frame-wise angular patterns from the physical sensor and replay them in simulation. This establishes sensor-level alignment by matching the simulated ray sampling process to the physical LiDAR.
For each valid return $\mathbf{p}_{k,i}=[x_{k,i},y_{k,i},z_{k,i}]^\top$ in scan $k$, after removing returns with \texttt{NaN} coordinates, we compute its azimuth and elevation as
\begin{equation}
(\vartheta_{k,i},\varphi_{k,i})
=
\Pi_{\mathrm{ang}}(\mathbf{p}_{k,i}),
\end{equation}
where $\Pi_{\mathrm{ang}}$ computes the azimuth as $\operatorname{mod}(\operatorname{atan2}(y_{k,i},x_{k,i}),2\pi)$ and the downward elevation as $\operatorname{atan2}(-z_{k,i}, \sqrt{x_{k,i}^{2}+y_{k,i}^{2}})$, so that $\varphi=0$ is horizontal and $\varphi=\pi/2$ points vertically downward.
From each scan with at least $N$ valid returns, we sample $N$ angular pairs without replacement to form the frame-wise pattern
\begin{equation}
\mathcal{A}_k
=
\left\{(\vartheta_{k,i},\varphi_{k,i})\right\}_{i=1}^{N},
\end{equation}
while scans with fewer valid returns are discarded. The resulting patterns are stored in acquisition order and replayed cyclically at the LiDAR update rate. Each angular sample defines a unit ray in the LiDAR frame,
\begin{equation}
\mathbf{d}_{k,i}
=
\begin{bmatrix}
\cos\varphi_{k,i}\cos\vartheta_{k,i} \\
\cos\varphi_{k,i}\sin\vartheta_{k,i} \\
-\sin\varphi_{k,i}
\end{bmatrix}.
\end{equation}
At simulation time $t$, let the valid intersection distances for ray $i$ be
\begin{equation}
\begin{aligned}
\Lambda_{t,i}
&=\left\{\lambda\in[r_{\min},r_{\max}]\;\middle|\right.\\
&\quad\left.\mathbf{p}^{W}_{t}+\lambda\mathbf{R}^{W}_{t}\mathbf{d}_{k_t,i}\in\mathcal{M}_t\right\}.
\end{aligned}
\end{equation}
The ray range is
\begin{equation}
\hat{r}_{t,i}=\begin{cases}
\displaystyle\min\Lambda_{t,i}, & \Lambda_{t,i}\neq\varnothing,\\[0.4em]
\mathrm{invalid}, & \Lambda_{t,i}=\varnothing.
\end{cases}
\end{equation}
where $k_t$ indexes the replayed pattern, $\mathbf{p}^{W}_{t}$ and $\mathbf{R}^{W}_{t}$ denote the LiDAR pose in the world frame, and $\mathcal{M}_t$ denotes the geometry queried by ray casting at time $t$. If the set of valid intersections is empty, we store an invalid/no-return marker for that ray. Following \cite{wang2025omni}, we use NVIDIA Warp \cite{macklin2022warp} to parallelize ray--mesh intersection queries across rays and environments. Moreover, \textbf{we incorporate dynamically updated robot geometry to model real-time self-occlusion.} Specifically, $\mathcal{M}_t$ comprises both the environment geometry and the collision meshes of the robot links extracted from the URDF, whose poses are updated at every LiDAR measurement using the corresponding rigid-body states. Since each ray retains its nearest valid intersection, robot links naturally occlude farther environmental surfaces without heuristic masking. Rays with no valid intersection retain the invalid/no-return marker and are mapped to the unobserved value during range-image encoding.

\subsection{Representation-Level Range-Image Encoding}
We encode both simulated and real LiDAR returns as spherical range images that serve as the exteroceptive input to the locomotion policy. Simulated Cartesian returns are reconstructed as
$\mathbf{p}^{\mathrm{sim}}_{t,i}=\hat{r}_{t,i}\mathbf{d}_{k_t,i}$,
while real returns are read directly from the onboard \texttt{PointCloud2} stream. Both domains then share the same encoding pipeline, establishing representation-level alignment between simulation and deployment. Since a single scan from a non-repetitive LiDAR may sparsely cover its field of view, we aggregate returns over a short temporal window $\Delta T=0.1\,\mathrm{s}$. Each scan is motion-deskewed, transformed into the current base frame using the calibrated LiDAR extrinsics, and compensated for the robot roll, pitch, and yaw at its acquisition time. This robot-centric aggregation improves angular coverage without global registration or persistent mapping. Let $\mathcal{P}_t=\{\mathbf{p}_{t,i}\}_{i=1}^{N_t}$ denote all points acquired in the current temporal window. After discarding invalid/no-return samples and points outside $\vartheta\in[0,2\pi)$ or $\varphi\in[0,\pi/2]$, we retain points satisfying $x\in[-1.25,1.5]\,\mathrm{m}$, $y\in[-0.75,0.75]\,\mathrm{m}$, and $r\le r_{\mathrm{enc}}=1.5\,\mathrm{m}$. Let $\mathcal{Q}_t=\{\bar{\mathbf{p}}_{t,j}\}_{j=1}^{J_t}\subseteq\mathcal{P}_t$ denote the retained points. For each retained point, we compute its range and angular coordinates as
\begin{equation}
    r_{t,j}
    =
    \left\|\bar{\mathbf{p}}_{t,j}\right\|_2,
    \qquad
    (\vartheta_{t,j},\varphi_{t,j})
    =
    \Pi_{\mathrm{ang}}(\bar{\mathbf{p}}_{t,j}),
\end{equation}
using the convention defined above. We uniformly discretize $\vartheta\in[0,2\pi)$ and $\varphi\in[0,\pi/2]$ into a $W\times H$ grid, with $(\vartheta_{\min},\vartheta_{\max})=(0,2\pi)$ and $(\varphi_{\min},\varphi_{\max})=(0,\pi/2)$:
\begin{equation}
\begin{aligned}
u_{t,j}
&=
\left\lfloor
\frac{\vartheta_{\max}-\vartheta_{t,j}}
{\vartheta_{\max}-\vartheta_{\min}}
(W-1)
\right\rfloor, \\
v_{t,j}
&=
\left\lfloor
\frac{\varphi_{\max}-\varphi_{t,j}}
{\varphi_{\max}-\varphi_{\min}}
(H-1)
\right\rfloor .
\end{aligned}
\end{equation}
Let
$\mathcal{C}_t(u,v)=\{j\mid(u_{t,j},v_{t,j})=(u,v)\}$.
The spherical range image
$\mathbf{I}_t\in\mathbb{R}^{W\times H}$ is defined as
\begin{equation}
\mathbf{I}_t[u,v]
=
\begin{cases}
\displaystyle
r_{\mathrm{enc}}^{-1}
\min_{j\in\mathcal{C}_t(u,v)} r_{t,j},
& \mathcal{C}_t(u,v)\neq\varnothing, \\[0.8em]
-1,
& \mathrm{otherwise}.
\end{cases}
\end{equation}
Each cell therefore stores the nearest normalized return, while invalid/no-return samples are mapped to the unobserved marker $-1$. Empty cells likewise receive $-1$, yielding a fixed-size representation independent of point count and ordering.

To improve transfer robustness, \textbf{we apply LiDAR-specific domain randomization (DR) during training} by perturbing the roll and pitch used for attitude compensation, individual ranges, and spatial cropping boundaries by $\epsilon_{\rho}$, $\epsilon_{\beta}$, $\epsilon_r$, and $\epsilon_b$, respectively. The entire pipeline is implemented with batched GPU operations, with per-cell nearest ranges computed via parallel scatter-min reduction. At deployment, randomization is disabled and the same deterministic encoder is applied directly to real measurements without online mapping or adaptation.
\subsection{End-to-End Locomotion Policy Learning}\label{sec:policy}
To isolate the effect of exteroceptive representation, we retain the Stage-1 policy architecture and training setup of Extreme Parkour \cite{cheng2024extreme} and change only its exteroceptive input to our spherical range image $\mathbf{I}_t$. The reward formulation, terrain curriculum, dynamics randomization, and PPO procedure are otherwise unchanged. No teacher--student distillation is used. The range image is encoded by an MLP $E_{\psi}$ as $\mathbf{z}^{\mathrm{ext}}_t=E_{\psi}\!\left(\operatorname{vec}(\mathbf{I}_t)\right)$.
The action issued at the preceding control step is denoted by $\mathbf{a}_{t-1}$. We provide it as an actor input, consistent with the policy interface in Fig.~\ref{fig:method}. Together with the proprioceptive observation $\mathbf{o}^{\mathrm{prop}}_t$, history latent $\mathbf{z}^{\mathrm{hist}}_t$, and command $\mathbf{c}_t$, the actor input is
\begin{equation}
\mathbf{o}^{\pi}_t
=
\left[
\mathbf{o}^{\mathrm{prop}}_t,
\mathbf{z}^{\mathrm{ext}}_t,
\mathbf{z}^{\mathrm{hist}}_t,
\mathbf{c}_t,
\mathbf{a}_{t-1}
\right].
\end{equation}
The actor outputs joint-position offsets tracked by a PD controller. Unlike privileged scandots, our range-image representation is available in both simulation and the real world. \textbf{This enables end-to-end reinforcement learning of perception and locomotion, allowing direct deployment without teacher--student distillation.}

\section{Experiments}
Our experiments address three questions: \textbf{Q1:}~Can the proposed LiDAR perception pipeline support learning challenging locomotion skills? \textbf{Q2:}~Does the range-image policy achieve stronger sim-to-real locomotion transfer? \textbf{Q3:}~Do our sensor-level simulation components: measured scan-pattern replay, self-occlusion modeling, and LiDAR domain randomization, improve sim-to-real transfer?

\begin{figure}[t]
    \centering
    \includegraphics[width=\linewidth]{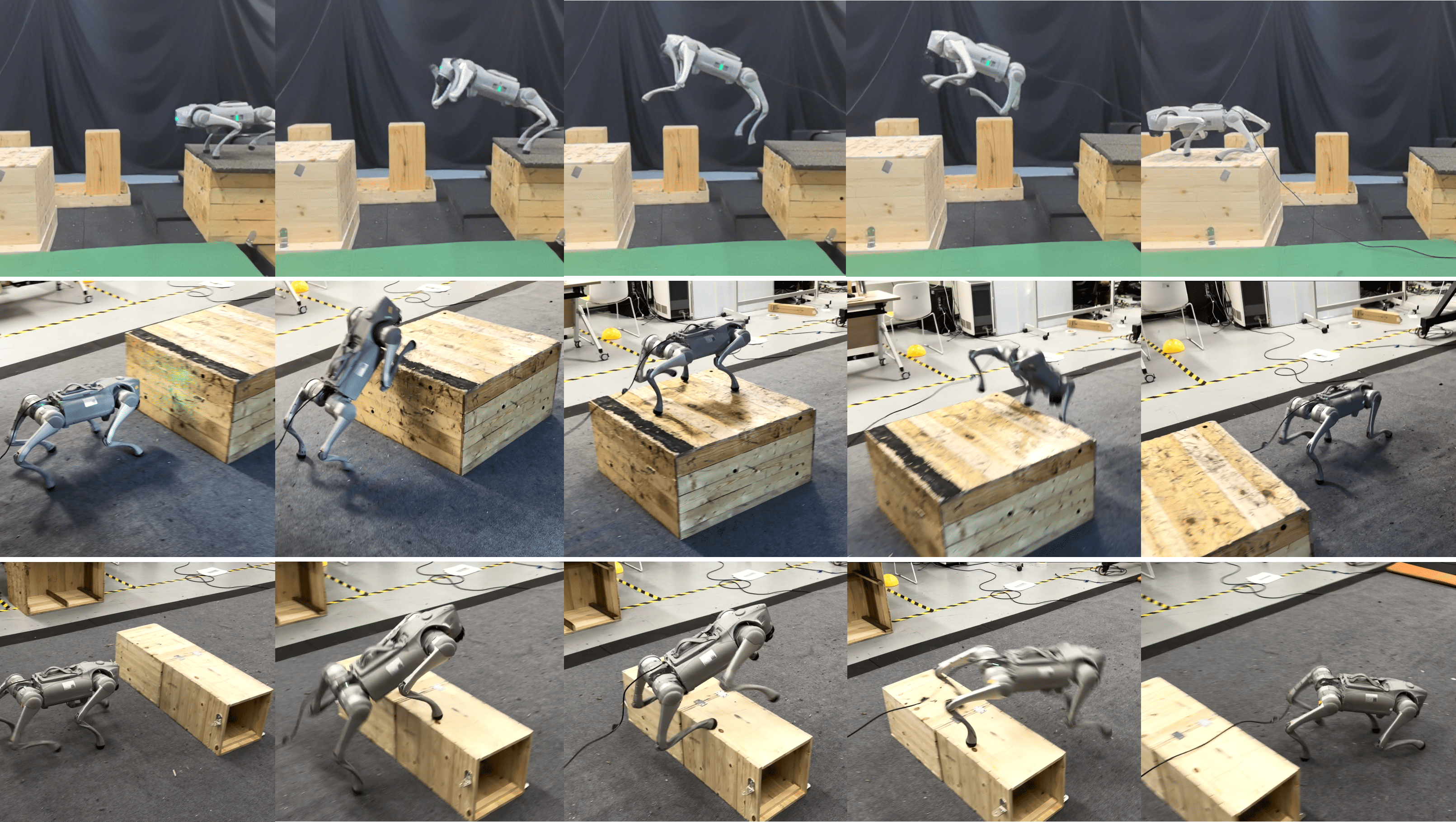}
    \caption{Real-world evaluation of our method on a Unitree Go2 with L2 LiDAR across three challenging terrain tasks: \textbf{Gap}, jumping across a 0.9\,m gap. \textbf{Step}, climbing onto a 0.45\,m high platform, and \textbf{Hurdle}, traversing over a 0.4\,m high obstacle.}
    \label{fig:snapshots}
    \vspace{-1em}
\end{figure}
\subsection{Experimental Setup}
We evaluate in simulation and on two Unitree Go2 robots equipped with Unitree L1 and L2 LiDARs, respectively. To enable direct comparison with Extreme Parkour \cite{cheng2024extreme}, we use its native gap, step, and hurdle terrain generators together with the same training and test terrain maps. The difficulty ranges are $0.1$--$0.9\,\mathrm{m}$ for gap width, $0.1$--$0.45\,\mathrm{m}$ for step height, and $0.1$--$0.4\,\mathrm{m}$ for hurdle height. For each method and terrain, we first select in simulation the highest difficulty $L$ at which the method achieves at least 80\% success, and then evaluate the same $L$ on hardware. Each terrain is tested in 20 trials. We follow the Extreme Parkour implementation for the depth baseline. For the elevation-map baseline, we use the stage-1 local height-map policy from Extreme Parkour in simulation. During hardware deployment, the elevation map is constructed using Point-LIO for odometry and the GridMap implementation released with PGTT \cite{ntagkas2025pgtt}. All LiDAR-based baselines and ablations use the L1 sensor. Only the explicitly labelled L2 row uses L2. Raw-LiDAR and PointNet baselines use the same temporal aggregation as our method, per-frame standardization, and a fixed number of normalized XYZ points. They apply no spatial cropping or range filtering, and PointNet receives XYZ coordinates directly. We ablate scan-pattern replay, self-occlusion, and LiDAR DR.

\subsection{Implementation Details}
We train $4096$ parallel Isaac Gym environments on one RTX~4090 GPU. The terrain curriculum, rewards, action space, PPO hyperparameters, control frequency, and PD gains follow the Stage-1 teacher of Extreme Parkour \cite{cheng2024extreme}. We replace its privileged scandots with a $20\times40$ range image encoded by an MLP with hidden dimensions $[128,64,32]$. The actor uses $[512,256,128]$ hidden units. We collect $1500$ real LiDAR scans to recover the frame-wise angular patterns, sampling $N=1000$ rays per L1 scan and $N=4000$ per L2 scan. The recovered patterns are replayed cyclically in acquisition order at each LiDAR update. The simulated update rate is randomized over $14$--$16\,\mathrm{Hz}$, and returns within the preceding $\Delta T=0.1\,\mathrm{s}$ are aggregated into each range image. Perception randomization is sampled per episode as
$\epsilon_{\rho},\epsilon_{\beta}\sim\mathcal{U}(-4^\circ,4^\circ)$,
$\epsilon_b\sim\mathcal{U}(-0.05,0.05)\,\mathrm{m}$, and
$\epsilon_r\sim\mathcal{U}(-0.05,0.05)\,\mathrm{m}$.
The policy requires no distillation, mapping, or online adaptation.
\begin{table}[t]
    \centering
    \caption{
        Traversal success rates (\%) in simulation and real deployment at
        the same difficulty level. Entries give Sim./Real success rates,
        followed by difficulty $L$. We select $L$ in simulation as the
        highest level with at least 80\% success and reuse it on hardware.
    }
    \label{tab:zero_shot_traversal}
    \small
    \setlength{\tabcolsep}{2pt}
    \renewcommand{\arraystretch}{1.08}
    \begin{tabular*}{\columnwidth}{@{\extracolsep{\fill}}p{0.42\columnwidth}ccc@{}}
        \toprule
        \textbf{Methods} & \multicolumn{3}{c}{\textbf{Terrain tasks}} \\
        \cmidrule(lr){2-4}
        & Gap & Step & Hurdle \\
        & \multicolumn{3}{c}{
          Sim./Real (\%) $\uparrow$ [difficulty $L$] $\uparrow$} \\
        \midrule

        \multicolumn{4}{@{}l@{}}{\underline{\textit{Baseline Methods}}} \\
        Extreme Parkour \cite{cheng2024extreme}
        & 85/40\% [9]
        & 95/35\% [9]
        & 100/75\% [9] \\
        Elevation map
        & 95/35\% [9]
        & 100/0\% [9]
        & 100/5\% [9] \\

        \midrule
        \multicolumn{4}{@{}l@{}}{\underline{\textit{LiDAR Representations}}} \\
        PointNet \cite{qi2017pointnet}
        & 80/45\% [4]
        & 90/10\% [9]
        & 85/45\% [9] \\
        Raw LiDAR data
        & 85/40\% [3]
        & 90/0\% [9]
        & 85/55\% [6] \\

        \midrule
        \multicolumn{4}{@{}l@{}}{\underline{\textit{Component Ablations}}} \\
        \shortstack[l]{w/o scan-pattern replay}
        & 90/20\% [9]
        & 90/10\% [9]
        & 100/50\% [9] \\
        \shortstack[l]{w/o LiDAR DR}
        & 95/50\% [9]
        & 95/40\% [9]
        & 100/80\% [9] \\
        \shortstack[l]{w/o self-occlusion modeling}
        & 90/60\% [9]
        & 95/50\% [9]
        & 100/95\% [9] \\

        \midrule
        \multicolumn{4}{@{}l@{}}{\underline{\textbf{\textit{Our Method}}}} \\
        L1 LiDAR
        & 90/60\% [9]
        & 90/55\% [9]
        & 100/100\% [9] \\
        L2 LiDAR
        & 95/65\% [9]
        & 90/50\% [9]
        & 100/95\% [9] \\

        \bottomrule
    \end{tabular*}
\end{table}

\subsection{Results}
\begin{figure*}[t]
    \centering
    \includegraphics[width=\textwidth]{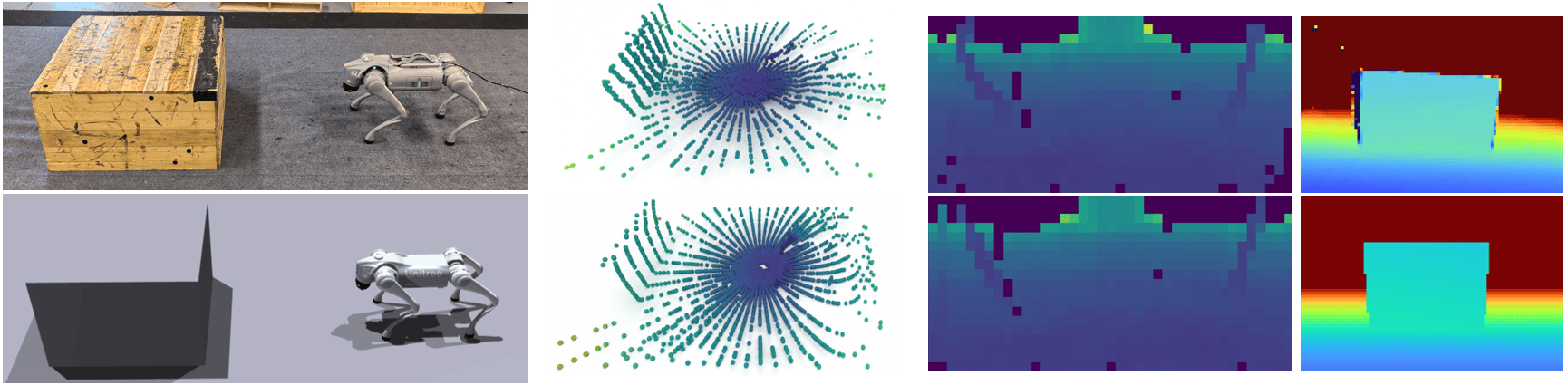}
    \caption{Qualitative comparison of real and simulated observations. The top row shows real-world data and the bottom row shows the corresponding simulated observations. From left to right: robot pose and scene state, corrected accumulated LiDAR point cloud, spherical range image, and forward-facing depth image rendered by the simulated D435i camera.}
    \label{fig:real-sim}
    \vspace{-1em}
\end{figure*}

\textbf{A1: The range-image pipeline enables learning of challenging locomotion skills.}
Table~\ref{tab:zero_shot_traversal} examines whether the proposed range-image observation supports learning challenging locomotion skills in simulation. The elevation-map baseline, which uses the privileged Stage-1 policy of Extreme Parkour, reaches the maximum difficulty $L=9$ on all three terrains and serves as an oracle reference. Its depth-based Stage-2 student also reaches $L=9$, with only modest loss after distillation. In contrast, Raw LiDAR reaches $L=3/9/6$ and PointNet $L=4/9/9$ on gap, step, and hurdle, respectively. We also observe less natural gaits and more frequent terrain contacts with these two policies. We attribute this to the difficulty of extracting control-relevant geometry from irregular, unstructured point sets. PointNet partially alleviates this issue, but still learns from an implicitly structured set representation. Our range image instead organizes LiDAR returns into a fixed spherical grid before policy learning, making local terrain geometry explicit and consistently accessible. Consequently, both L1 and L2 policies reach $L=9$ on all three terrains with 90--100\% simulated success.

\noindent \textbf{A2: The range-image policy achieves stronger sim-to-real locomotion transfer.}
Although the depth and elevation-map baselines both reach $L=9$ in simulation, their performance drops sharply on hardware at the same terrain levels (0.9\,m gap, 0.45\,m step, and 0.4\,m hurdle). The depth policy achieves only 40/35/75\% success. Across hardware trials, depth measurements varied substantially with illumination and robot motion, with unstable invalid pixels around object boundaries. Failed step trials showed that these artifacts distorted the perceived platform edge and were frequently accompanied by premature or delayed front-leg lifting. We attribute these failures partly to this observation shift, absent from the simulated depth used to train the distilled student. The elevation-map policy, despite the strongest simulation performance, drops to 35/0/5\%. Its Stage-1 policy is trained with complete privileged height maps, whereas real maps must be reconstructed from partial LiDAR observations. Obstacle occlusion often produced incomplete local height maps on the step and hurdle tasks, while accumulated Point-LIO drift further misaligned the extracted height maps. The gap geometry remained visible, consistent with its higher success rate. Our range-image policies retain substantially more simulated performance. L1 and L2 achieve 60/55/100\% and 65/50/95\% real-world success, averaging 71.7\% and 70.0\%, versus 50.0\% for depth and 13.3\% for elevation maps. Their sim-to-real retentions are 76.8\% and 73.7\%, compared with 53.6\% and 13.6\%. We attribute this smaller gap to sensor- and representation-level alignment design in our pipeline. Figure~\ref{fig:real-sim} further visualizes this cross-domain consistency. Raw-LiDAR and PointNet policies also exhibit unnatural motions and frequent body or joint contacts on hardware, consistent with their weaker simulation performance.

\noindent \textbf{A3: Sensor-level simulation components improve transfer.}
We also isolate sensing components for sim-to-real transfer while keeping the range-image representation fixed. All ablations reach $L=9$ with 90--100\% simulated success, yet differ substantially on hardware. The largest degradation occurs without scan-pattern replay: replacing the recovered patterns with a fixed uniform pattern reduces mean real-world success from 71.7\% to 26.7\%, despite nearly unchanged simulation performance. Thus, matching only the ray count and field of view is insufficient, reproducing the physical sensor's frame-wise angular sampling is important for transfer. Removing LiDAR-specific domain randomization reduces mean success to 56.7\%, showing that range and preprocessing perturbations help absorb residual sensing mismatch. Self-occlusion modeling has a smaller but consistent effect, with mean success decreasing to 68.3\% and 5-point drops on both step and hurdle. These ablations reveal a gap that simulation alone cannot expose: the policies remain equally capable in simulation, while sensing fidelity determines how much of that capability transfers to hardware.

\section{Conclusion}
We presented a LiDAR perception pipeline for end-to-end locomotion that aligns sensing and representation between simulation and reality. The proposed spherical range image provides a structured LiDAR observation that supports learning challenging locomotion skills, while scan-pattern replay, self-occlusion modeling, and shared encoding reduce perceptual discrepancies across domains. Experiments show stronger hardware transfer than other baselines, and ablations identify scan-pattern replay as the most influential component, with LiDAR-specific domain randomization and self-occlusion modeling providing additional gains.

\clearpage

\section{Compliance with Ethical Standards}
This study did not involve human or animal subjects and therefore
did not require ethical approval.

\section{Acknowledgments}
The authors have no financial or nonfinancial interests to disclose.

\bibliographystyle{IEEEbib}
\bibliography{refs}

\end{document}